\documentclass[10pt,twocolumn,letterpaper]{article}

\usepackage{iccv}
\usepackage{times}
\usepackage{ifpdf}
\usepackage{epsfig,epstopdf}
\usepackage{graphicx}
\usepackage{amsmath}
\usepackage{amssymb}

\usepackage[pagebackref=true,breaklinks=true,letterpaper=true,colorlinks,bookmarks=false]{hyperref}

\iccvfinalcopy 

\def\iccvPaperID{4760} 

\ificcvfinal\fi
\begin{document}

\title{
Effects of Blur and Deblurring to Visual Object Tracking
}

\author{Qing Guo$^{1,2}$,\, Wei Feng$^{1,2}$\thanks{Corresponding author. Tel: (+86)-22-27406538. This work is supported by NSFC 61671325, 61572354, 61672376.},\, Zhihao Chen$^{1,2}$,\, Ruijun Gao$^{1,2}$,\, Liang Wan$^{1,2}$,\, Song Wang$^{1,2,3}$\\
	$^1$ Colledge of Intelligence and Computing, Tianjin University, Tianjin 300350, China\\
	$^2$ Key Research Center for Surface Monitoring and Analysis of Cultural Relics (SMARC),\\ State Administration of Cultural Heritage, China\\
	$^3$ Department of Computer Science and Engineering, University of South Carolina, Columbia, SC 29208, USA\\
	{\tt\small \{tsingqguo,wfeng,zhouce,ruihuang,lwan\}@tju.edu.cn, songwang@cec.sc.edu}
}

\maketitle

\begin{abstract}

%
%
%
%
%
%
%
%

Intuitively, motion blur may hurt the performance of visual object tracking.
However, we lack quantitative evaluation of a tracker's robustness to different levels of motion blur. 
Meanwhile, while image-deblurring methods can produce visually clearer videos for pleasing human eyes, it is unknown whether visual object tracking can benefit from image deblurring or not.
In this paper, we address these two problems by constructing a Blurred Video Tracking benchmark, which contains a variety of videos with different levels of motion blurs, as well as ground-truth tracking results for evaluating trackers. 
We extensively evaluate 23 trackers on this benchmark and observe several new interesting results. 
Specifically, we find that light blur may improve the performance of many trackers, but heavy blur always hurts the tracking performance. 
We also find that image deblurring may help to improve tracking performance on heavily-blurred videos but hurt the performance on lightly-blurred videos. 
According to these observations, we propose a new GAN-based scheme to improve the tracker's robustness to motion blurs. 
In this scheme, a fine-tuned discriminator is used as an adaptive assessor to selectively deblur frames during tracking process. 
We use this scheme to successfully improve the accuracy and robustness of 6 trackers.

\end{abstract}

\section{Introduction}

\begin{figure}[t]
	\begin{center}
		\includegraphics[width=0.98\linewidth]{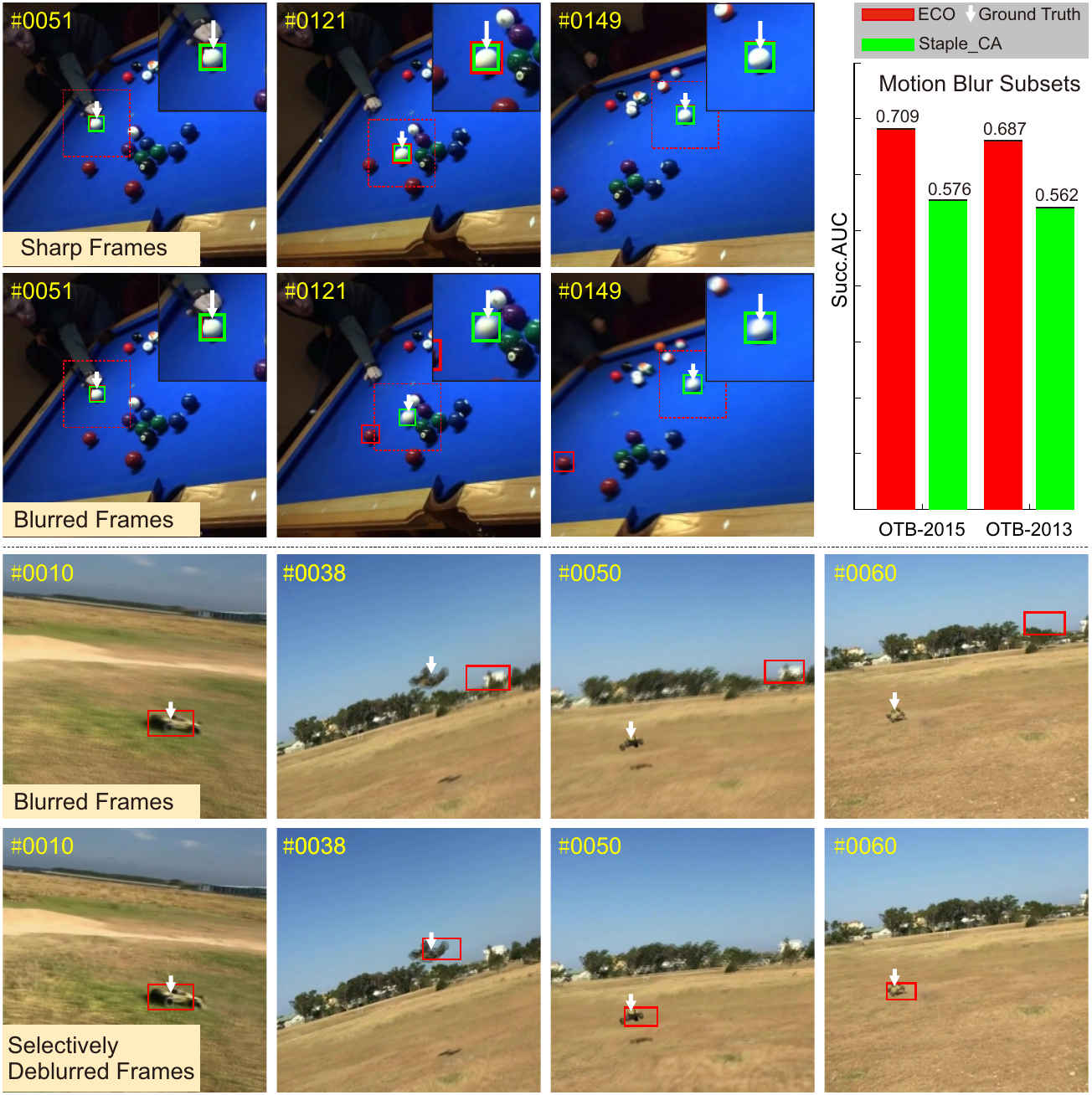}\vspace{-1em}
	\end{center}
	\caption{\small{Results of ECO~\cite{Danelljan16ECO} and Staple\_CA~\cite{Mueller17,Bertinetto16} on sharp, blurred or deblurred videos captured from one scene. The first subfigure shows that ECO locates the target accurately on sharp frames while losing it on blurred ones. In contrast, Staple\_CA can capture the ball in both cases. Such situation is not considered by existing benchmarks, e.g. OTB~\cite{Wu13,Wu15} in which ECO has much higher accuracy than Staple\_CA. The bottom subfigure presents that ECO misses the target on blurred frames while locating it accurately when we selectively deblur the frames. }}
	\label{fig:motivation}\vspace{-1.1em}
\end{figure}

Motion blur caused by camera shake and object movement not only reduces the visual perception quality, but also may severely degrade the performance of video analysis tasks, e.g. single object tracking~\cite{Wu15}. In recent years, numerous tracking benchmarks are proposed to evaluate how well current trackers can handle motion blur by comparing their accuracy on videos containing blurred frames~\cite{Wu13,Wu15,Liang2015TIP,Fan2019LaSOT}. However, such benchmarks do not exclude the influence of other possible interferences, e.g. the limitation of the algorithms of the time, thus leads to incomplete conclusion of a tracker, since the accuracy may be underestimated due to other issues. In addition, with current blur-related tracking benchmark, we cannot quantitatively evaluate trackers' robustness to different levels of motion blur, thus cannot support deep exploration of the way that motion blur affects tracking performance.

As shown in Fig.~\ref{fig:motivation}, given a sharp video and its blurred version for the same scene, ECO~\cite{Danelljan16ECO} can locate the white billiard ball accurately on the sharp video but fail to do so when motion blur happens. In contrast, Staple\_CA~\cite{Bertinetto16,Mueller17} tracks the ball accurately on both videos. This situation can not be thoroughly evaluated on blurred videos captured under different scenes. For example, OTB benchmark~\cite{Wu13,Wu15} shows that ECO has much higher accuracy than Staple\_CA on its motion blur subsets, which certainly does not consider the above situation. A comprehensive benchmark that fairly measures the blur robustness of trackers is necessary and will encourage the development of blur-robust trackers. 

A naive solution for blur-robust tracking is to first deblur the frames of a video and then apply trackers on the deblurred video. However, it is known that such naive deblurring strategy may introduce ringing artifacts, due to the Gibbs phenomenon that hurts the features of raw frames and fails the tracking easily~\cite{Ding2016TCSVT,Ma2016TIP,Jin2005CVPR,Wu2011ICCV,Wu2011AVC}. Instead of direct deblurring, many recent blur-aware trackers add different kinds of blur to the target template, forming an augmented template set, then locate the target at following frames by matching candidates with all of the blur augmented templates~\cite{Wu2011ICCV,Ma2016TIP}. Although such trackers are effective but they have high memory and computing cost. Besides, how to do effective blur augmentation is unknown.

Note, the negative effects of deblurring to visual tracking are concluded mainly based on early deblurring algorithms. Recently, numerous successful deblurring methods have been developed via deep learning, with significantly improved performance, fewer artifact noises and much faster speed~\cite{Kupyn2018CVPR,Nah2017CVPR,Noroozi2017,Sun2015CVPR,Xu2014NIPS,Sun2015CVPR}. But, whether they are helpful for visual object tracking still remains questionable.

In this paper, we aim to analyze the effects of motion blur and deblurring methods to current trackers, and explore an effective way of using existing deep deblurring to achieve blur-robust tracking. Our main contributions are three-fold:

\begin{itemize}
	\item{
		We construct a Blurred Video Tracking~(BVT) benchmark with a dataset containing 500 videos for 100 scenes. Each scene consists of 5 videos having different levels of motion blurs. 	We use three metrics to evaluate the accuracy and blur robustness of trackers.}
	
	\item{
		We extensively evaluate 23 trackers on the BVT benchmark and find that the light motion blur improves most of the trackers, while the heavy blur hurt their accuracy significantly. We also find that deblurring methods can improve the tracking performance on heavily-blurred videos, while having negative effects to the ones with light blur.
	}
	
	\item{
		We propose a new GAN-based tracking scheme that adopts the fine-tuned discriminator of DeblurGAN as an adaptive blur assessor to selectively deblur frames during the tracking process and improve the accuracy of 6 state-of-the-art trackers. 
	}
\end{itemize}

\section{Related Work}

\subsection{Tracking benchmarks}

In recent years, numerous tracking benchmarks have been proposed for general performance evaluation or specific issues~\cite{Smeulders2014PAMI,Wu13,Wu15,Liang2015TIP,Kristan15,Kristan16,Mueller2016ECCV,Li2016PAMI,Kristan2017ICCVW,Mueller2018ECCV,Fan2019LaSOT,huang2018got}. 
The OTB~\cite{Wu13,Wu15}, ALOV++~\cite{Smeulders2014PAMI}, VOT~\cite{Kristan16,Kristan15,Kristan2017ICCVW}, TrackingNet~\cite{Mueller2018ECCV}, LaSOT~\cite{Fan2019LaSOT} , and GOT-10K~\cite{huang2018got} benchmarks provide unified platforms to compare state-of-the-art trackers.  
More recent ones, e.g. TrackingNet, LaSOT and GOT-10K, contain a large scale of videos and cover a wide range of classes, which will make training a high performance deep learning based trackers available. 
Other benchmarks focus on specific applications or problems. 
For example, the NfS~\cite{Galoogahi2017ICCV} benchmark consists of 100 high frame rate videos and analyze the influence of appearance variation to deep and correlation filter-based trackers respectively.

Among these benchmarks, the OTB-2013~\cite{Wu13}, OTB-2015~\cite{Wu15}, TC-128~\cite{Liang2015TIP}, and LaSOT~\cite{Fan2019LaSOT} datasets contain motion blur subsets that can be used to evaluate the ability of trackers to handle the motion blur.
Nevertheless, the evaluation results are incomplete, since other interference that also affects the tracking accuracy is not excluded.

A better solution is to compare trackers on the videos that are captured at the same scene but have different levels of motion blur to see if the tracker can obtain the same performance. 
In this paper, we construct a dataset for motion blur evaluation by averaging the frames on high frame rate videos with different ranges, thus generate testing videos having the same content with different levels of motion blur. 
By doing this, we are able to score the robustness of trackers and help study the effects of motion blur.

\subsection{Motion blur-aware trackers}
Numerous works have studied the relationship between the motion blur and the object tracking~\cite{Ma2016TIP,Seibold2017CVIU,Jin2005CVPR,Dai2006ICIP,Mei2008CVPR,Wu2011AVC,Wu2011ICCV}.
Jin et al.~\cite{Jin2005CVPR} have observed that matching between blurred images help realize effective object tracking.
In \cite{Dai2006ICIP,Mei2008CVPR,Wu2011AVC}, how to estimate the blur kernel accurately during object tracking is carefully studies.
Ma et al.~\cite{Ma2016TIP} and Wu et al.~\cite{Wu2011ICCV} propose to integrate the visual object tracking with the motion blur problem through sparse representation and realize blur robust trackers. 

Above works are studied according to the observation that deblurring methods can introduce negative effects to frames and corrupt the features.
However, deblurring methods have achieved great progress in recent years. 
Whether the latest works are helpful for object tracking remains questionable.
A recent work~\cite{Seibold2017CVIU} finds that motion blur is helpful and provides additional motion information of the target.  
However, this work does not discuss the effects of different levels of the motion blur to object tracking.

\subsection{Other state-of-the-art trackers}
Latest tracking works focus on construct powerful appearance models to realize high performance tracking.
We can coarsely split recent works into three categories including correlation filter~(CF) based~\cite{Li18,Danelljan16ECO,Bertinetto16,AL17-csrdcf,DSARCF_TIP2019,Galoogahi17}, classification\&updating based~\cite{MDNet16,Song2018CVPR,Jung2018ECCV} and Siamese network or matching based~\cite{Bertinetto16-2,Guo17_ICCV,Zhu2018ECCV,Wang2018CVPR,Wang2019CVPR,Fan2019CVPR} trackers.

Although these trackers have achieved great performance improvement on benchmarks, there is no specific benchmark that can evaluate their ability to handle different levels of motion blur.


\subsection{GAN based methods}
Generative adversarial networks~(GANs)~\cite{Goodfellow2014} is to train two competitors, i.e. the discriminator and the generator.  
The generator is to produce fake samples that can fool the discriminator.
The discriminator is to separate fake samples from real ones.
With recent studies~\cite{Arjovsky2017,Gulrajani2017} to alleviate training problems of GAN~\cite{Salimans2016}, it has helped achieve great progress in areas of deblurring~\cite{Kupyn2018CVPR}, superresolution~\cite{Ledig2017CVPR}  and image painting~\cite{Yeh2017CVPR,Dolhansky_2018_CVPR} and other related problems.  

Nevertheless, most of the GAN-based methods just regard the discriminator as a part of loss function to train the generator and discard it during testing time.
In this paper, we find that the discriminator trained for DeblurGAN~\cite{Kupyn2018CVPR} can score the blur level of motion blur and help realize selective deblurring for blur-robust tracking.
%

\section{Blurred Video Tracking~(BVT) Benchmark}
\label{sec:brb}

\subsection{Dataset}

\begin{figure}[t]
	\begin{center}
		\includegraphics[width=0.98\linewidth]{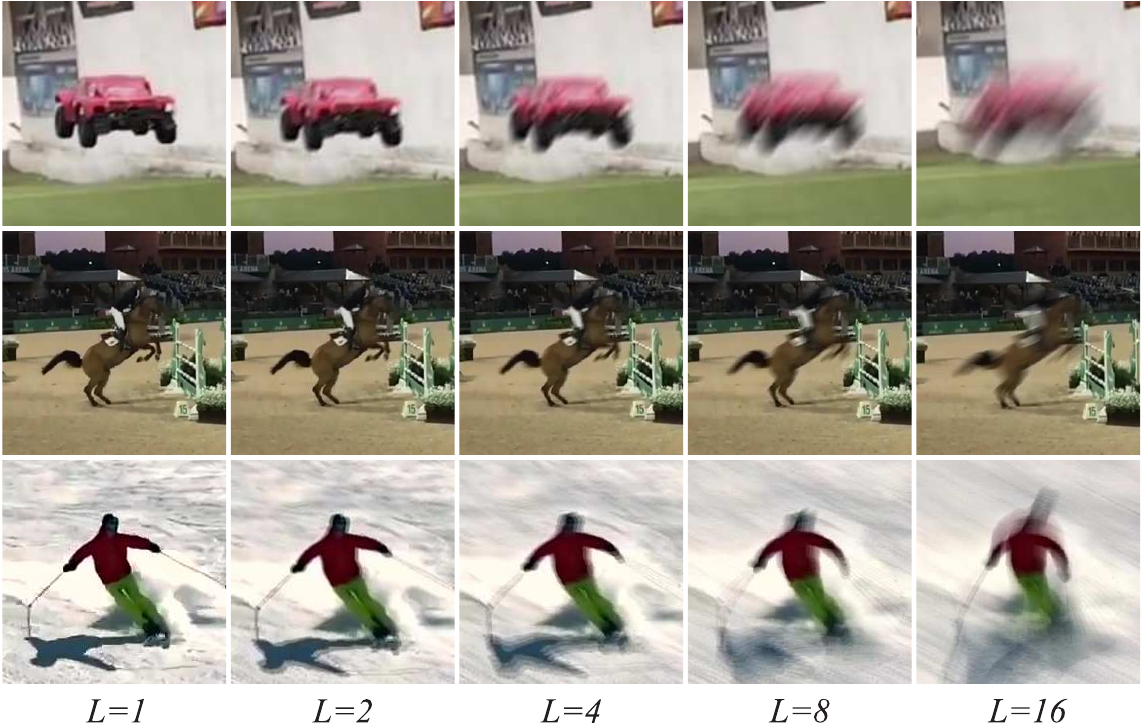}\vspace{-1em}
	\end{center}
	\caption{\small{Examples of frames that are blurred in 5 levels. `$L=1$' represents the blurred video contains raw frames that are captured at 240~fps and have the least serious blur.}}
	\label{fig:blurlevels}\vspace{-1.1em}
\end{figure}

Galoogahi et al.~\cite{Galoogahi2017ICCV} proposed the NfS dataset that consists of 100 videos captured at 240~fps. 
Since frames in such high frame rate videos are sharp, we can generate realistic motion blur with different levels by averaging these sharp frames, as done in deblurring methods~\cite{Nah2017CVPR,Noroozi2017} . 
Given a video $\mathcal{V}=\{\mathbf{I}_t\}_1^T$ in the NfS dataset, we produce a blurred one $\tilde{\mathcal{V}}^L=\{\tilde{\mathbf{I}}_t^L\}_1^{\tilde{T}}$ each frame of which is the average of $L$ successive frames of $\mathcal{V}=\{\mathbf{I}_t\}_1^T$, i.e. $\tilde{\mathbf{I}}_t^L = \mathrm{avg}(\{\mathbf{I}_t,...,\mathbf{I}_{t+L-1}\})$. 
The length $L$ decides the level of motion blur, that is, a larger $L$ leads to more serious blur. The ground truth of the target in $\tilde{\mathbf{I}}^L_t$ is set as the average of annotations of medium frames in $\{\mathbf{I}_t,...,\mathbf{I}_{t+L-1}\} $. 

The blurred video, i.e. $\tilde{\mathcal{V}}^L=\{\tilde{\mathbf{I}}_t^L\}_1^T$, are still at the high frame rate, and the difference between neighbor frames is small. This will affect the blur robustness evaluation since a simple tracker can also obtain high accuracy on high frame rate videos~\cite{Galoogahi2017ICCV}.
We then temporally sample $\tilde{\mathcal{V}}^L$ at every 8 frames and obtain a new video denoted as $\bar{\mathcal{V}}^L=\{\bar{\mathbf{I}}_t^L\}_1^{\bar{T}}$ whose frame rate is 30~fps. Note, to avoid the initialized target template containing motion blur, we borrow the first frame from the high frame rate video, i.e. $\mathcal{V}$, and set it as the first frame of $\bar{\mathcal{V}}^L$.  

Following the above setup, for each video in the NfS dataset, we generate 5 blurred videos by setting $L=1,2,4,8,16$.
`$L=1$' represents the video $\bar{\mathcal{V}}^1$ contains raw frames with the least serious blur. 
All these videos make up a new dataset denoted as $\mathcal{S}$ that contains 500 videos and consists of 5 subsets, i.e. $\mathcal{S}=\{\mathcal{S}^L|L=1,2,4,8,16\}$, corresponding to 5 different levels of motion blur.
Fig.~\ref{fig:blurlevels} shows 3 cases of blurred frames in 5 different levels.
Clearly, through temporal averaging on high frame rate frames, we obtain realistic blurred videos in which the blur is directly related to the object and camera motion pattern.
When the camera is fixed, the object is heavily-blurred while the background is still sharp.
Such results are not easily achieved by using synthetic technologies.

\subsection{Metrics}

 \begin{figure*}[t]
	\begin{center}
		\includegraphics[width=0.95\linewidth]{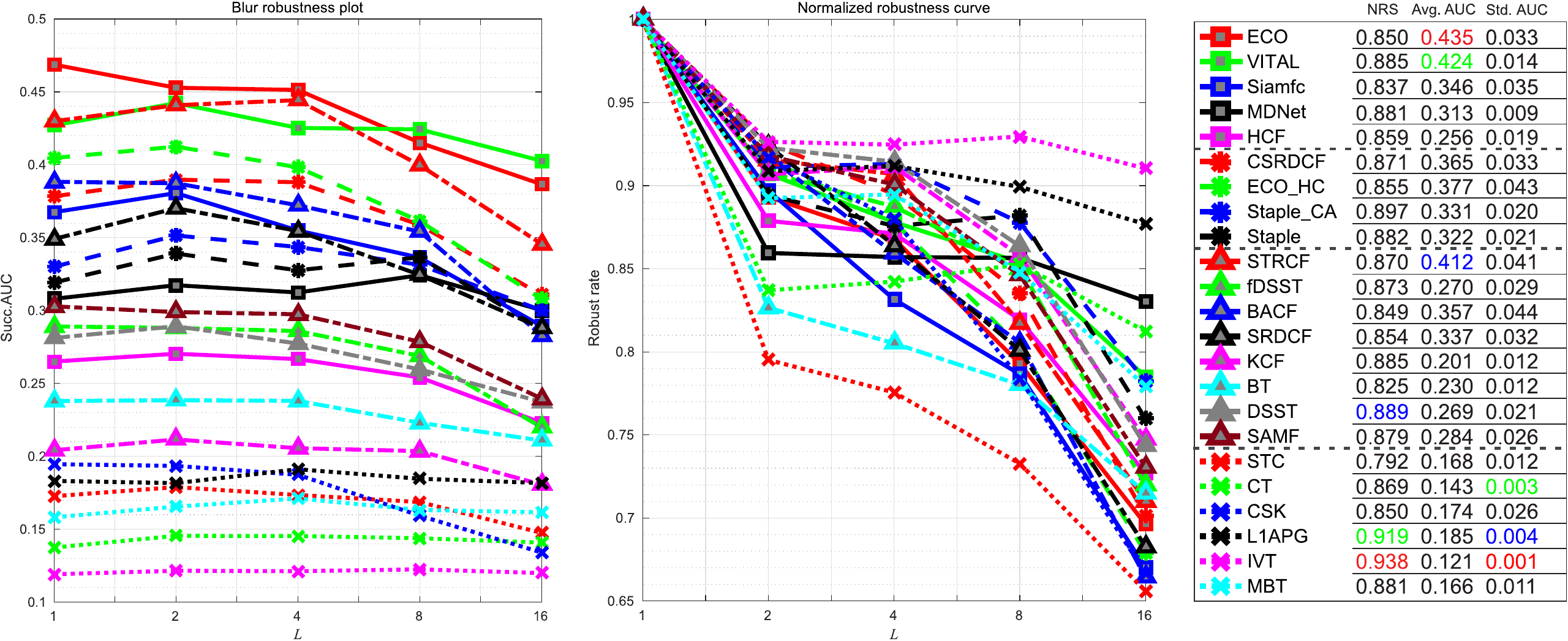}\vspace{-1em}
	\end{center}
	\caption{\small{Evaluation results of 23 trackers on the BVT benchmark. 
			The left subfigure shows the blur robustness plot of each tracker.   
			The medium subfigure presents normalized robustness curves of all trackers. 
			The right subfigure displays the normalized robustness score~(NRS), average AUC and its standard variation on 5 subsets of each tracker respectively. }}
	\label{fig:eval_results}\vspace{-1.1em}
\end{figure*}

We set three metrics for the blur robustness evaluation based on the success metric defined in \cite{Wu15}. 
Specifically, we first calculate the intersection over union~(IoU) between predicted and annotated bounding boxes at each frame of a subset $\mathcal{S}^L$.
We then draw a success plot which presents the percentage of bounding boxes whose IoU is larger than given thresholds.   
The area under curve~(AUC) of the success plot is used to compare different trackers on the subset $\mathcal{S}^L$ and is denoted as $A^L$.
Given a tracker, we can obtain 5 AUC scores for the 5 subsets and draw a blur robustness plot with the X-axis representing different subsets and Y-axis being AUC scores.
We can rank compared trackers according to the average and standard variance of AUC scores respectively.
The average of 5 AUC scores measures the absolute accuracy of a tracker on different blurred videos while the standard variance represents the robustness.

In addition, we propose a new metric named as normalized robustness score to make the blur robustness be independent to the accuracy.
Specifically, we first evaluate a tracker on the sharp video subset, i.e. $\mathcal{S}^1$, and obtain a set of frames denoted as $\mathcal{I}_\mathrm{succ}^{1}$ on which the tracker can locate the target accurately while the IoU is larger than 0.5.
Note, each frame of $\mathcal{I}_\mathrm{succ}^1$ has corresponding blurred versions on other subsets and we denote them as $\mathcal{I}_\mathrm{succ}^L$ where $L>1$.
We then run the tracker on other blurred subsets, i.e. $\mathcal{S}^{\{2,4,8,16\}}$, and calculate the average IoUs on $\mathcal{I}_\mathrm{succ}^{\{2,4,8,16\}}$ respectively. 
We finally get a normalized vector by
\begin{equation}\label{eq:nrc}
\mathbf{u}=\frac{[u_1,u_2,u_4,u_8,u_{16}]}{u_1},
\end{equation}
where $u_L$ is the average IoU on $\mathcal{I}_\mathrm{succ}^L$, and $\mathbf{u}$ corresponds to a normalized robustness curve~(NRC). The average of all elements in $\mathbf{u}$ is denoted as the normalized robustness score~(NRS). 
If the NRS of a tracker approximates to 1, it means that the tracker is not affected by the motion blur and can still locate the target on blurred versions of $\mathcal{I}_\mathrm{succ}^1$.
 
\section{Evaluation Results}
 With the proposed BVT benchmark, we evaluate 23 trackers and analyze their blur robustness.
 Meanwhile, we use two state-of-the-art deep deblurring methods to handle the blurred subsets of the BVT benchmark and discuss how these methods can help improve tracking performance.
 
\subsection{Effects of blur to tracking}
\label{subset:effect_blur}
\begin{figure}[t]
	\begin{center}
		\includegraphics[width=0.95\linewidth]{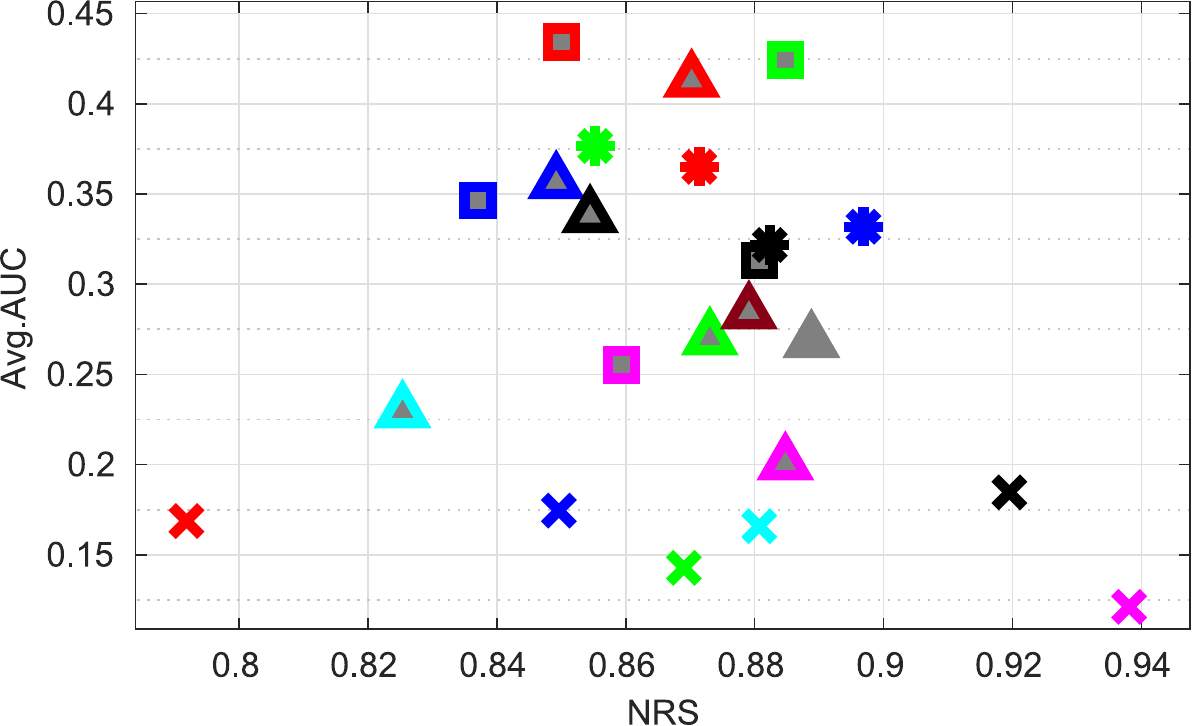}\vspace{-1em}
	\end{center}
	\caption{\small{
	Normalized robustness score~(NRS) and average AUC of 23 trackers.  The legend is the same with that of Fig.~\ref{fig:eval_results}.
}}
	\label{fig:acc-robust}\vspace{-1.1em}
\end{figure}

 {\bf Trackers.}  
 We evaluate 23 trackers on the proposed benchmark and categorize them into 4 classes according to representations they used: trackers using intensity based features~\footnote{Here, the intensity based features consist of the template used by IVT~\cite{Ross07}, L1APG~\cite{Bao12}, CSK~\cite{Henriques12} and STC~\cite{Zhang14-stc}, and haar-like features used by CT~\cite{Zhang12ECCV}.}, i.e. IVT~\cite{Ross07}, L1APG~\cite{Bao12}, CT~\cite{Zhang12ECCV}, CSK~\cite{Henriques12}, STC~\cite{Zhang14-stc} and MBT~\cite{Ma2016TIP}, trackers based on HoG features, i.e. BT~\cite{Wang15}, DSST~\cite{Danelljan14}, KCF~\cite{Henriques15}, SAMF~\cite{Li14-samf}, SRDCF~\cite{Danelljan15}, fDSST~\cite{Danelljan17}, BACF~\cite{Galoogahi17} and STRCF~\cite{Li18}, trackers with deep features, i.e. HCF~\cite{Ma15}, ECO~\cite{Danelljan16ECO}, MDNet~\cite{Nam16}, Siamfc~\cite{Bertinetto16-2} and VITAL~\cite{Song2018CVPR}, trackers using mixed features, i.e. Staple~\cite{Bertinetto16}, Staple\_CA~\cite{Mueller17}, ECO\_HC~\cite{Danelljan16ECO} and CSRDCF~\cite{AL17-csrdcf}. 
 
 {\bf Overall results.}
 We present the evaluation results on Fig.~\ref{fig:eval_results} and \ref{fig:acc-robust}.
 In general, the accuracy of trackers decreases with the increase of the motion blur level. 
 In terms of the average AUC, ECO achieves the highest accuracy on the BVT benchmark while VITAL~\cite{Song2018CVPR} is in the second place, since these trackers employ deep features as object representations and are equipped with sophisticatedly designed online learning strategies.
 Among trackers base on hand-crafted features, STRCF~\cite{Li18}, ECO\_HC~\cite{Danelljan16ECO} , and CSRDCF~\cite{AL17-csrdcf} are in the first, second, and third places respectively according to the average AUC. 
Moreover, these trackers are better than MDNet~\cite{Nam16} and HCF~\cite{Ma15} that use deep features.
The trackers using intensity-based features have much lower accuracy than others due to the less discriminative power.
 
It terms of the robustness evaluation, we observe that trackers using intensity-based features are generally more robust to motion blur, since they obtain similar accuracy on both heavily-blurred and sharp videos.
Specifically, IVT~\cite{Ross07} has the highest normalized robustness score~(NRS) and smallest standard variance of AUC while L1APG~\cite{Bao12} gets the second high NRS.
STC~\cite{Zhang14-stc} and CT~\cite{Zhang12ECCV} have bad NRSs while their standard variations of AUC are very small.

 As shown in Fig.~\ref{fig:acc-robust}, considering both average AUC and NRS, we find that Staple\_CA~\cite{Mueller17} achieves well balance between the accuracy and blur robustness.
 Although VITAL~\cite{Song2018CVPR} is slightly worse than ECO~\cite{Danelljan16ECO} on the average AUC, it has much higher NRS than ECO.
 According to blur robustness plots, we find that the rank of trackers has great difference on 5 subsets. 
 For example,  VITAL~\cite{Song2018CVPR} obtains smaller AUC score than ECO~\cite{Danelljan16ECO} and STRCF~\cite{Li18} on $\mathcal{S}^1$ while being the best one on $\mathcal{S}^{16}$. 
 We can find similar results on BACF~\cite{Galoogahi2017ICCV}, CSRDCF~\cite{AL17-csrdcf}, DSST~\cite{Danelljan14} and CSK~\cite{Henriques12}. 

In summary, we have following observations:
\textit{
Simply comparing trackers on a single subset is not enough to conclude their abilities to handle motion blur.
The accuracy and blur robustness of trackers are dependent on features they used.
Trackers using intensity-based features obtain low accuracy while usually being robust to motion blur.
Deep features help track accurately but are somehow sensitive to severe blur.
It is necessary to explore possible combination strategies to take both advantages.
}

\begin{figure}[t]
 	\begin{center}
 		\includegraphics[width=0.95\linewidth]{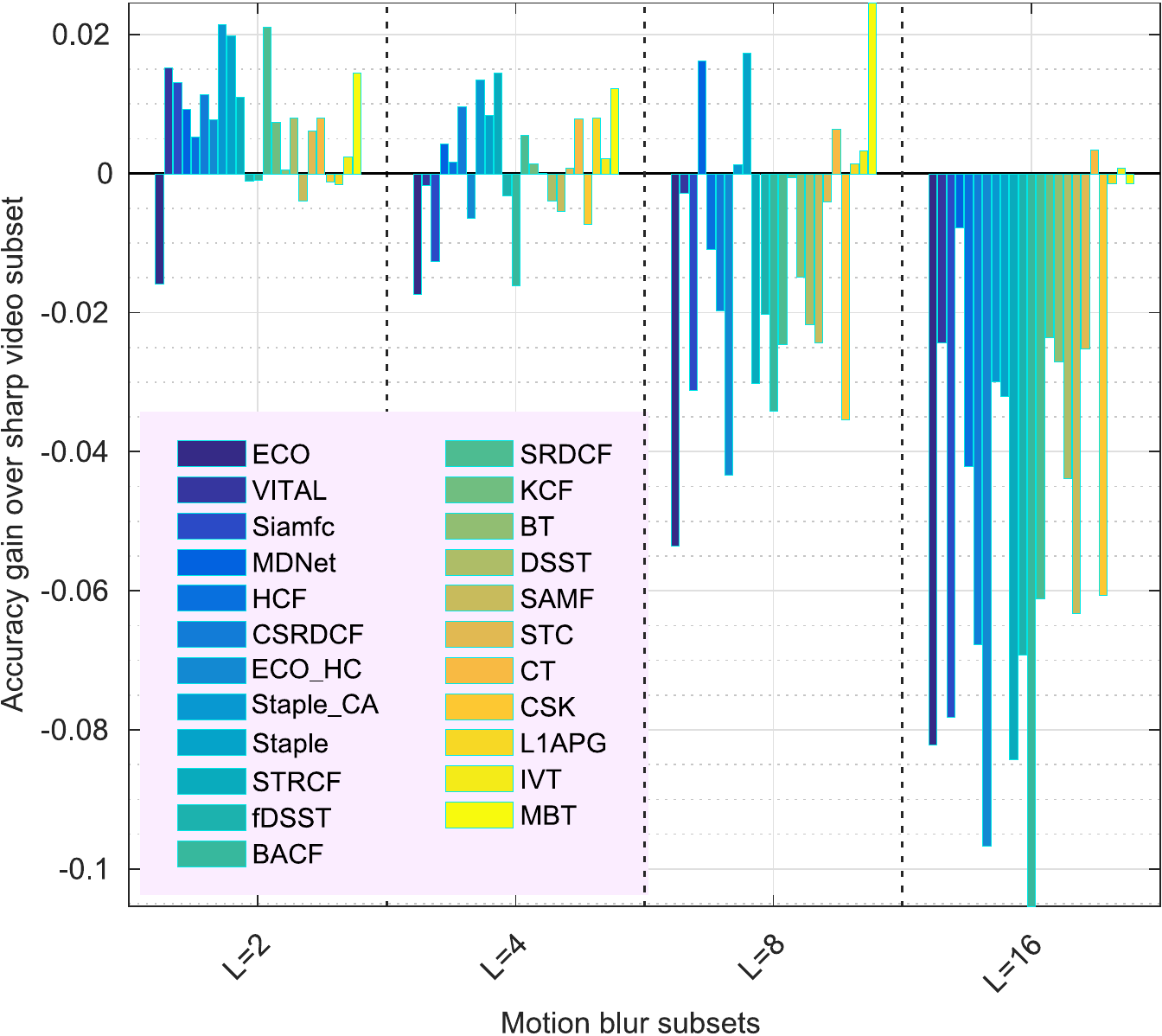}\vspace{-1em}
 	\end{center}
 	\caption{\small{
	AUC gains of blurred subsets, i.e. $\mathcal{S}^{\{2,4,8,16\}}$, over sharp video subset, i.e. $\mathcal{S}^1$ for all compared trackers. 
 	}}
 	\label{fig:diff-perf}\vspace{-1.1em}
 \end{figure}

 {\bf Benefits of light motion blur.}
According to blur robustness plots shown in Fig.~\ref{fig:eval_results}, a lot of trackers obtain higher AUC on lightly-blurred subsets, e.g. $\mathcal{S}^2$ and $\mathcal{S}^4$, than on $\mathcal{S}^1$, which infers that the light motion blur has positive effects on tracking performance. 
To better understand this observation, for each tracker, we calculate the AUC gain of blurred subsets, i.e. $\mathcal{S}^{\{2,4,8,16\}}$, over the sharp version, i.e. $\mathcal{S}^1$, through $G^{L}=A^{L}-A^1$ where $G^L>0$ means a tracker has higher accuracy on $\mathcal{S}^L$ than on $\mathcal{S}^1$.  
As shown in Fig.~\ref{fig:diff-perf}, on the lightly-blurred subsets, i.e. $\mathcal{S}^2$ and $\mathcal{S}^4$, there are 17 and 14 trackers that have positive gains. 
Such numbers reduce to 7 and 2 on heavily-blurred subsets, i.e. $\mathcal{S}^2$ and $\mathcal{S}^4$, respectively.
Hence, light motion blur does help most of the compared trackers obtain higher accuracy. 
This is because the lightly-blurred videos generated by averaging neighbor high rate frames contain more effective information for separating the target from the background.
 
For some specific methods, we find that ECO using deep features always obtains negative gains on all subsets with gradually enlarging magnitude.
We have similar observations on VITAL and Siamfc, although they obtain higher AUCs on $\mathcal{S}^1$.
In contrast, trackers with intensity-based features, e.g. IVT and CT, have positive gains on all subsets, which further demonstrate the importance of features in handling motion blur.
We also show that the motion blur-aware tracker, i.e. MBT~\cite{Ma2016TIP}, achieves positive gains on all subsets except $\mathcal{S}^16$ and has the highest gain on $\mathcal{S}^8$.
 
In summary, we have following observations:
\textit{Light motion blur helps most of the trackers achieve higher accuracy while heavy blur significantly reduces the performance of almost all trackers.}
\subsection{Effects of deblurring to tracking}
\label{subset:effect_deblur}

 \begin{figure*}[t]
	\begin{center}
		\includegraphics[width=0.95\linewidth]{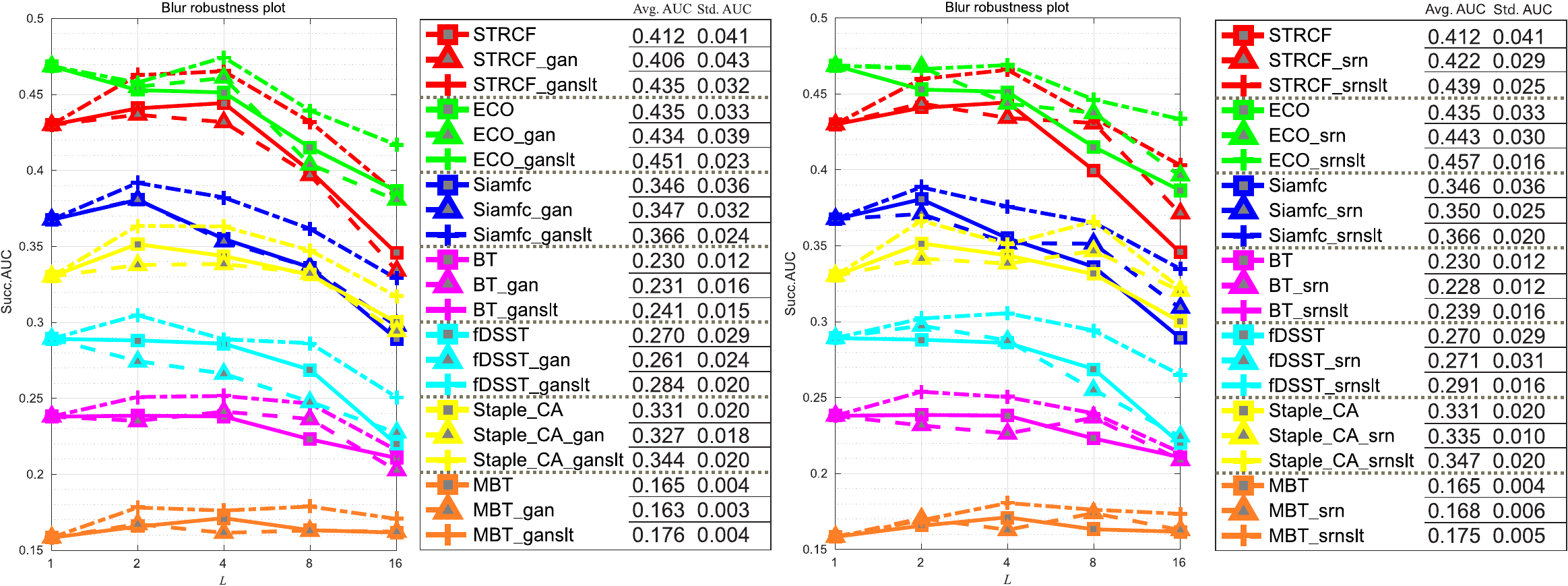}\vspace{-1em}
	\end{center}
	\caption{\small{
	Evaluation results of 7 typical trackers and their four variants.
	DeblurGAN~\cite{Kupyn2018CVPR} and Scale-recurrent network~(SRN)~\cite{Tao2018CVPR} are used to cope with the blurred frames respectively.
	`*\_gan' and `*\_srn' denote trackers deblurring each frame via DeblurGAN and SRN respectively. `*\_ganslt' and `*\_srnslt' are methods that selectively deblur frames according to the localization error of trackers.
	}}
	\label{fig:deblur_results}\vspace{-1.1em}
\end{figure*}

In following, we will study whether state-of-the-art deep deblurring methods could help improve the accuracy of trackers under the motion blur.

{\bf Methods.} 
Early deblurring methods run slowly and are not suitable for real-time tracking.
We select two deep deblurring methods, i.e. DeblurGAN~\cite{Kupyn2018CVPR} and SRN~\cite{Tao2018CVPR}, that run much faster via the GPU\footnote{DeblurGAN takes average 0.05~s to deblur search regions that are about 5~times larger than targets.} and achieve state-of-the-art deblurring performance. 
Given a tracker, we use a deblurring method to get two variants. 
The first one is to deblur all frames before tracking and we name it as the full deburring based method.
The second one is to selectively deblur frames during the tracking process according to center localization errors, i.e. the distance between predicted bounding boxes and ground truth.

With two deblurring methods, we get four variants for each tracker and denote them as `*\_gan', `*\_srn' for full deblurring based ones, and `*\_ganslt', `*\_srnslt' for selective deblurring based methods respectively, where `*' represents the name of a tracker.
We test these variants on four blurred video subsets, i.e. $\mathcal{S}^{\{2,4,8,16\}}$.

To achieve comprehensive study, we select 7 representative trackers including the ones that achieve best accuracy on the BVT benchmark, i.e. STRCF~\cite{Li18} and ECO~\cite{Danelljan16ECO}, the Siamese network based tracker, i.e. SiamFC~\cite{Bertinetto16-2}, CF trackers using hand-crafted features, i.e. fDSST~\cite{Danelljan17} and Staple\_CA~\cite{Bertinetto16,Mueller17}, a typical classification based tracker, i.e. BT~\cite{Wang15} and a motion blur-aware tracker, i.e. MBT~\cite{Ma2016TIP}.  

{\bf Cons of full deburring.}
As shown in Fig.~\ref{fig:deblur_results}, when we deblur all frames during tracking process via DeblurGAN, we get lower accuracy than using blurred frames at most of the time. 
The performance decline decreases as the motion blur level being serve.
For example, the AUC of fDSST\_gan is much smaller than that of fDSST on $\mathcal{S}^{2,4,8}$ while becomes slightly better on $\mathcal{S}^{16}$.
Such observation encourages that we should perform deblurring on heavily-blurred frames and pass the ones containing light blur, when we use DeblurGAN to improve the blur robustness.

In terms of the SRN method, by deburring all frames, it slightly improves most of the trackers.
Specifically, STRCF\_srn and ECO\_srn achieve 2.4\% and 1.8\% relative improvement over STRCF and ECO while performance gains on other trackers are very small and even negative.      
Similar with DeblurGAN, SRN helps trackers get higher improvement on heavily-blurred videos while making their accuracy drop on videos having light blur.
For example, STRCF\_srn has similar or even worse AUC score than STRCF on $\mathcal{S}^2$ and $\mathcal{S}^{4}$ while obtaining great improvement on $\mathcal{S}^8$ and $\mathcal{S}^{16}$. 
We have similar observations on the BT and Siamfc. 

In summary, we have following observations:
\textit{
State-of-the-art deep deblurring methods, i.e. DeblurGAN~\cite{Kupyn2018CVPR} and SRN~\cite{Tao2018CVPR}, usually result in tracking accuracy decreasing on lightly-blurred videos while having positive effects on the ones containing heavy motion blur.   
}

{\bf Pros of selective deblurring.}
 \begin{figure}[t]
	\begin{center}
		\includegraphics[width=0.98\linewidth]{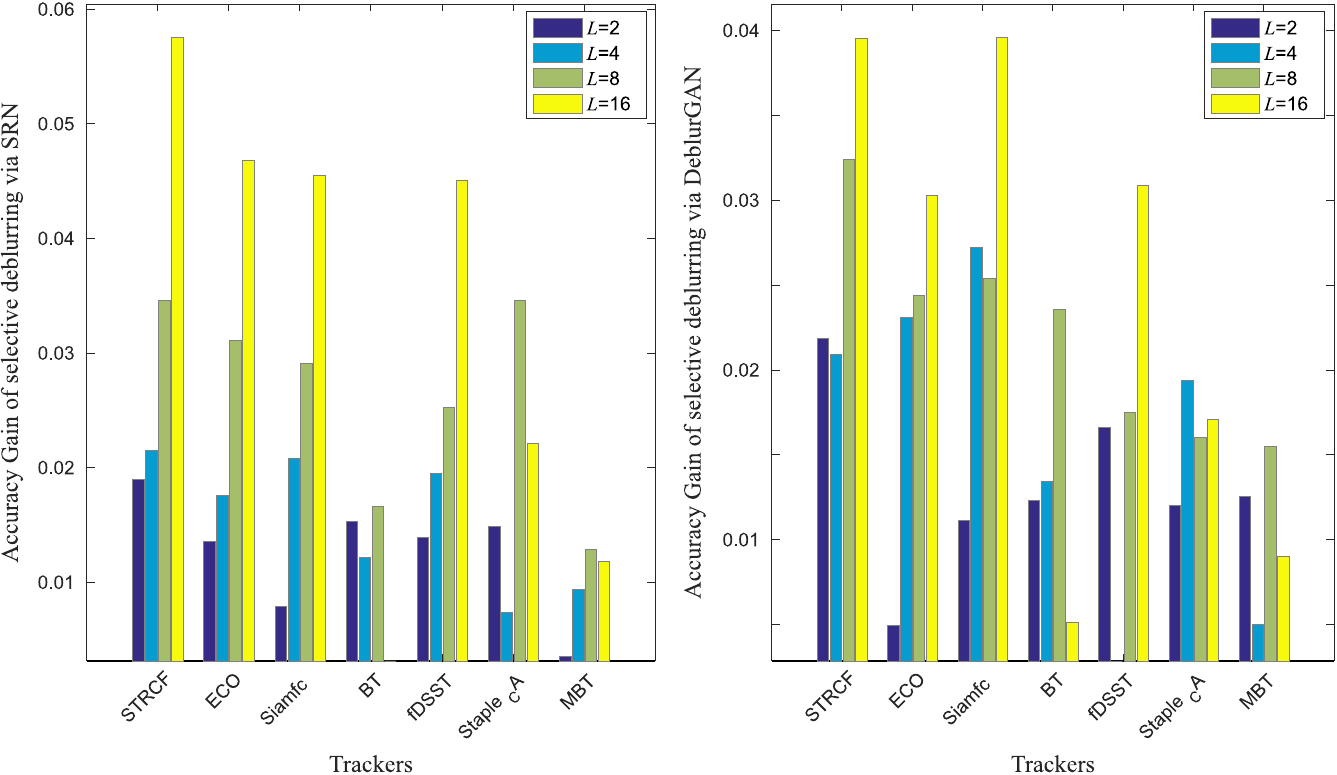}\vspace{-1em}
	\end{center}
	\caption{\small{
			AUC gains of selective deblurring based trackers, i.e. `*\_ganslt' and `*\_srnslt', over original ones on blurred video subsets, i.e. $\mathcal{S}^{\{2,4,8,16\}}$.
	}}
	\label{fig:gain_sltdeblur}\vspace{-1.1em}
\end{figure}
According to observations in Section.~\ref{subset:effect_blur} and \ref{subset:effect_deblur}, selective deblurring should help improve tracking performance. 
To validate this assumption, we selectively deblur an incoming frame according to localization errors during the tracking process.
Specifically, for the incoming frame $t$, we first use DeblurGAN or SRN to handle it and obtain a deblurred image. 
We then predict the target position according to raw and deblurred frames respectively and obtain two bounding boxes whose center localization errors are calculated according to the ground truth.
The result with higher precision is saved as the final output.
We name above method as `*\_ganslt' or `*\_srnslt'.

Fig.~\ref{fig:deblur_results} shows that selective deblurring via DeblurGAN and SRN improves the tracking performance of all trackers significantly.
%
%
Furthermore, we notice that selective debluring based methods generally have higher gain over the original versions on heavily-blurred videos than on light ones.
%
%
As shown in Fig.~\ref{fig:deblur_results}, the performance improvements of STRCF\_*slt, ECO\_*slt, Siamfc\_*slt and fDSST\_*slt w.r.t. their original versions gradually increase and reach their maximum on $\mathcal{S}^{16}$.
Other trackers, e.g. BT, Staple\_CA and MBT, have similar trend while achieving the highest gain on $\mathcal{S}^{8}$.

In summary, we have following observations:
\textit{
Selective deblurring improves tracking performance significantly. 
Accuracy gains incrementally increase with the growing motion blur level and generally reach the maximum at the most heavily-blurred video subset.
}
%

\section{Blur-Robust Tracking via DeblurGAN-D}

\subsection{DeblurGAN-D as blur assessor}
\begin{figure}[t]
	\begin{center}
		\includegraphics[width=0.98\linewidth]{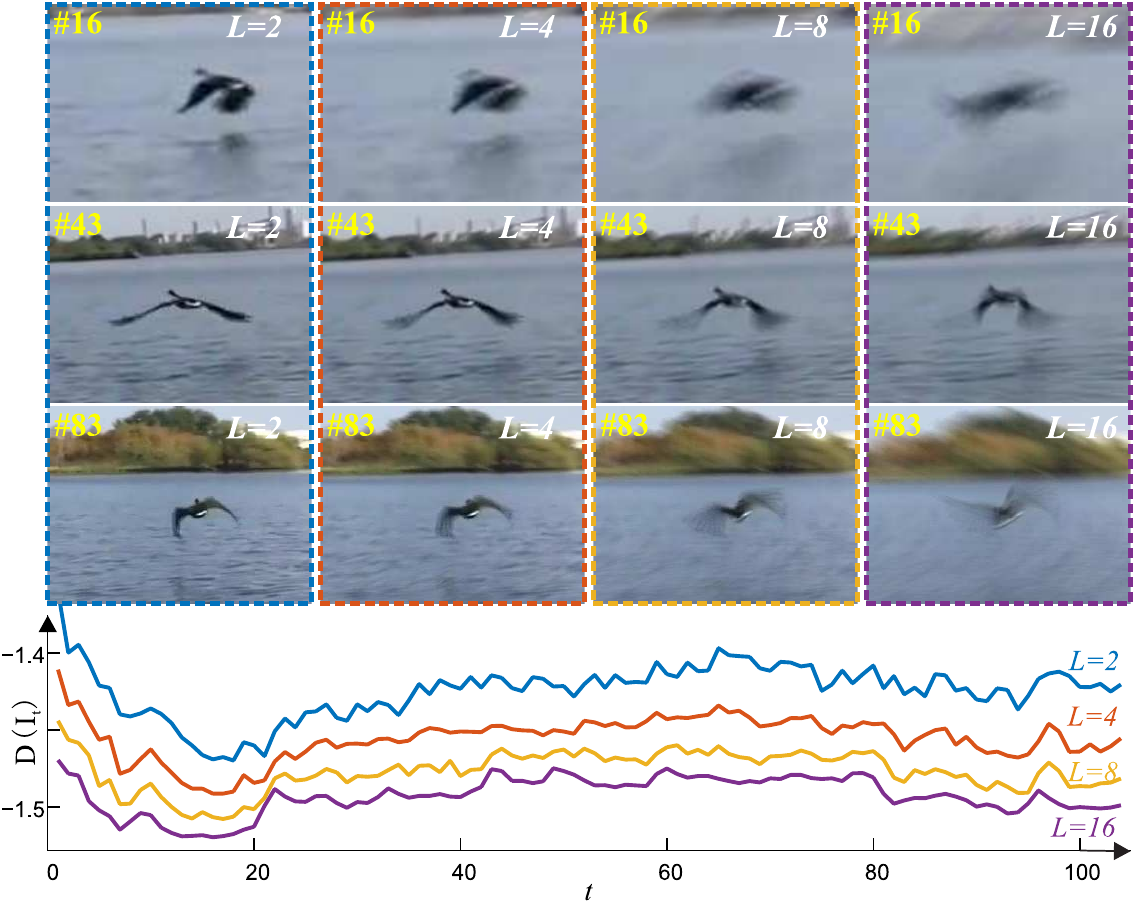}\vspace{-1em}
	\end{center}
	\caption{\small{
	Outputs of the discriminator of DeblurGAN, i.e. $\mathrm{D}(\cdot)$, on bird sequences that contain 4 levels of motion blur.
	}}
	\label{fig:disc_cases}\vspace{-1.1em}
\end{figure}

DeblurGAN~\cite{Kupyn2018CVPR} uses the critic network as the discriminator~($\mathrm{D}$) to output scores of sharp and restored images and calculate their Wasserstein distance as the loss to train the generator~($\mathrm{G}$) and the discriminator itself.
$\mathrm{D}$ only works at the training process and is discarded at testing time. 
In the training stage, $\mathrm{G}$ outputs deblurred images whose quality is gradually improved.
We can regard these images as blurred ones having different blur levels.
From the view of training $\mathrm{D}$, it is tuned to distinguish between sharp images and the ones generated by $\mathrm{G}$, which have different blur levels.
As a result, the discriminator has the ability to make a distinction between sharp and blurred images.

As shown in Fig.~\ref{fig:disc_cases}, we calculate discriminator outputs of frames in four videos that have different blur levels. 
Clearly, the heavily-blurred video, i.e. $L=16$, has the smallest value while the sharp one, i.e. $L=2$, has the highest score.
Hence, the discriminator of DeblurGAN is able to score the blur levels of frames and will help decide when we should do deblur during the tracking process. 

\subsection{Fine-tuning DeblurGAN-D}
\label{subsec:fine-tuning}

\begin{figure}[t]
	\begin{center}
		\includegraphics[width=0.98\linewidth]{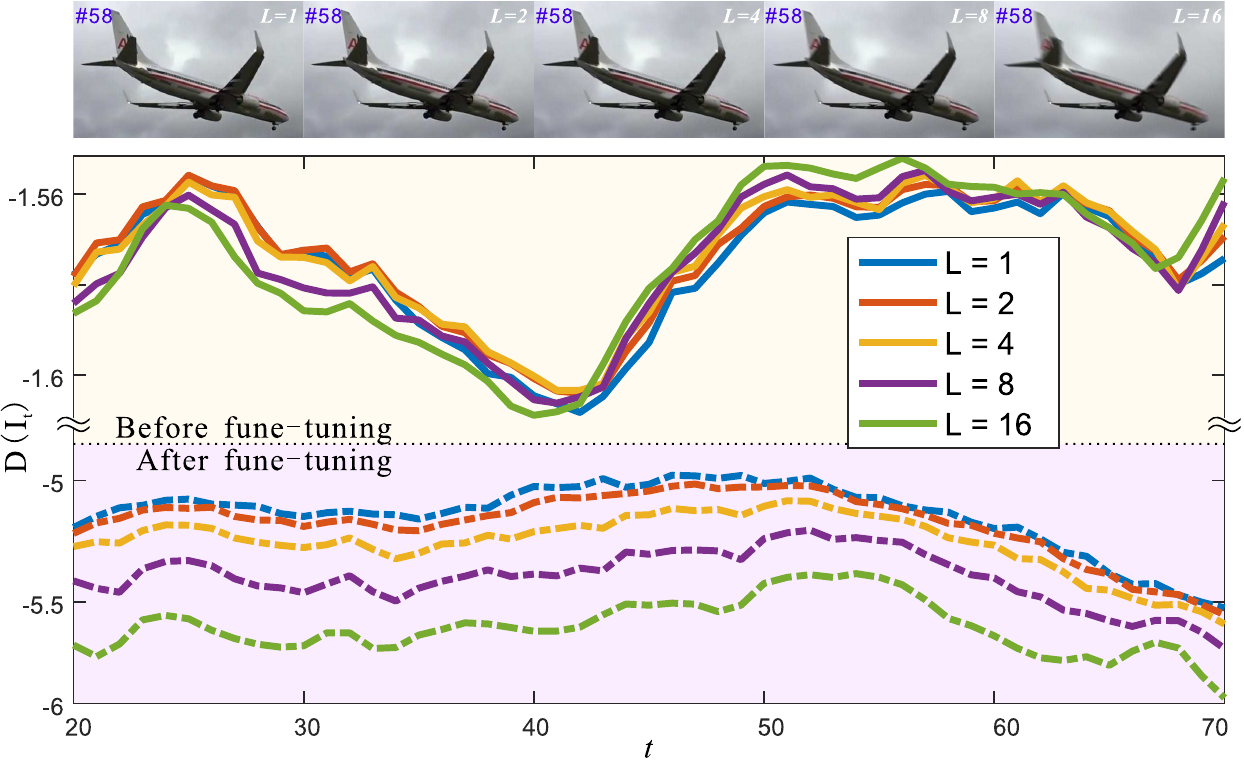}\vspace{-1em}
	\end{center}
	\caption{\small{
			Comparing the fine-tuned discriminator with the original one on airplane sequences.
	}}
	\label{fig:disc_case2}\vspace{-1.1em}
\end{figure}

Although we have shown DeblurGAN-D can score the blur degree of a frame, it easily fails and cannot discriminate motion blur degrees when their visual difference is small. 
As shown in Fig.~\ref{fig:disc_case2}, DeblurGAN-D cannot rank the blur degree of frames properly. 
This is because DeblurGAN-D is originally designed to compare the sharp and deblurred images, which has a gap to the task of assessing blur degrees.

To alleviate above problem, we propose to fine-tune DeblurGAN-D with blur \& deblur image pairs. 
Specifically, we select 20 scenes including 80 blurred videos from the dataset of the BRB and obtain 32304 frames.
Each scene contains 4 videos corresponding to 4 blur degrees respectively.   
We use the generator of DeblurGAN to deblur these frames and get 32304 blur \& deblur image pairs.
Using these pairs as training data, we particularly fine-tune the discriminator via the same adversarial loss of DeblurGAN with the fixed generator.

As shown in Fig.~\ref{fig:disc_case2}, compared with the original discriminator, the fine-tuned one can not only sort blur degrees properly but also reflect the distance between different motion blurs.
In practice, we calculate the discriminator difference between blurred and deblurred frames, i.e. $\mathrm{D}(\hat{\mathbf{I}}_t)-\mathrm{D}(\mathbf{I}_t)$, to avoid the influence of non-blur information in the image, where $\hat{\mathbf{I}}_t$ is the deblurred $\mathbf{I}_t$. 
A larger difference corresponds to a heavy motion blur of $\mathbf{I}_t$.

Note, we can also use blur \& sharp image pairs to train $\mathrm{D}$.
However, in real applications, sharp images are not given and we have to take extra cost to collect suitable images for fine-tuning.
In contrast, the proposed strategy does not need extra data and is also suitable for other deblurring methods. 
Please find more visualization results in supplementary material.   

\subsection{Selective deblurring for blur-robust tracking}

Given a video $\mathcal{V}=\{\mathbf{I}_t\}^{T}_1$, we formulate a tracker within Bayesian framework in which the maximum a posterior estimation of the target state at frame $t$ , i.e. a bounding box $\mathbf{b}_t$, is computed by 
\begin{equation}\label{eq:map_track}
P(\mathbf{b}_t|\mathcal{I}_t) = \sum_{s_t\in\{0,1\}} {P(\mathbf{b}_t|\mathcal{I}_t,s_t)P(s_t|\mathcal{I}_t)},
\end{equation}
where $\mathcal{I}_t=\{\mathbf{I}_1,...,\mathbf{I}_t\}$ is the set of observed frames, and $s_t$ is a selector that can be 1 and 0 representing to use deblurred and raw $\mathbf{I}_t$ to estimate $\mathbf{b}_t$ respectively. 
$P(s_t|\mathcal{I}_t)$ is used to estimate $s_t$ via observed frames and calculated by
\begin{equation}\label{eq:selector_generator}
P(s_t|\mathcal{I}_t)=\alpha_\mathrm{s} P(\mathbf{I}_t|s_t)\sum_{s_{t-1}\in\{0,1\}}{P(s_t|s_{t-1})P(s_{t-1}|\mathcal{I}_{t-1})},
\end{equation}
where $\alpha_\mathrm{s}$ is a normalization factor, $P(s_t|s_{t-1})$ is a motion model for the selector to consider historical selection results, and $P(\mathbf{I}_t|s_t)$ measures the necessity to deblur $\mathbf{I}_t$
\begin{equation}\label{eq:necess_deblur}
P(\mathbf{I}_t|s_t)\propto|\mathrm{D}(\hat{\mathbf{I}}_t)-\mathrm{D}(\mathbf{I}_t)|,
\end{equation}
where $\hat{\mathbf{I}}_t$ is the deblurred $\mathbf{I}_t$. 
Instead of directly using $\mathrm{D}(\mathbf{I}_t)$ for $P(\mathbf{I}_t|s_t)$, we calculate the difference between $\mathrm{D}(\hat{\mathbf{I}}_t)$ and $\mathrm{D}(\mathbf{I}_t)$ to remove the influence of non-blur information. 

In Eq.~(\ref{eq:map_track}),  the posterior probability of $\mathbf{b}_t$ being the target, given the selector $s_t$ and previous frames, i.e.$P(\mathbf{b}_t|\mathcal{I}_t,s_t)$ can be rewritten as
\begin{equation}\label{eq:map_loc}
\left\{
\begin{array}{lr}
 \alpha P(\hat{\mathbf{I}}_t|\mathbf{b}_t)\sum\limits_{\mathbf{b}_{t-1}}{P(\mathbf{b}_t|\mathbf{b}_{t-1})P(\mathbf{b}_{t-1}|\mathcal{I}_{t-1})}, \mathrm{if}\ s_t=1 &\\\\
  \alpha P(\mathbf{I}_t|\mathbf{b}_t)\sum\limits_{\mathbf{b}_{t-1}}{P(\mathbf{b}_t|\mathbf{b}_{t-1})P(\mathbf{b}_{t-1}|\mathcal{I}_{t-1})}, \mathrm{if}\ s_t=0 &
\end{array}
\right.  
\end{equation}
where $\alpha$ is a normalization factor, $P(\hat{\mathbf{I}}_t|\mathbf{b}_t)$ and $P(\mathbf{I}_t|\mathbf{b}_t)$ are observation models that compute the likelihood of $\mathbf{b}_t$ belonging to the target with inputs being $\hat{\mathbf{I}}_t$ and $\mathbf{I}_t$ respectively, and $P(\mathbf{b}_t|\mathbf{b}_{t-1})$ represents the motion model.

For an existing tracker, we can use its observation and motion models to calculate $P(\mathbf{b}_t|\mathcal{I}_t,s_t)$ via Eq.~(\ref{eq:map_loc}) and locate the target by solving
\begin{equation}\label{eq:track}
\mathbf{b}_t^* = \mathrm{arg}\max_{\mathbf{b}_t}P(\mathbf{b}_t|\mathcal{I}_t).
\end{equation}

In practice, given a tracker and an incoming frame, we crop a search region and deblur it with the DeblurGAN-G.
We then obtain 2 bounding boxes and their object likelihoods by feeding the tracker with raw and deblurred search regions.
When $|\mathrm{D}(\hat{\mathbf{I}}_t)-\mathrm{D}(\mathbf{I}_t)|>\theta$ where $\theta=2.5$ for all trackers, the search region is heavily-blurred and the bounding box of deblurred search region is saved as final result. Otherwise, the one having largest object likelihood is saved.   
Currently, we set $P(s_t|s_{t-1})$ as a discrete uniform distribution that ignores the historical selection results and will discuss other possible ways in the future.

We can equip extensive existing trackers with the proposed scheme. 
In following, we will validate the scheme on 7 trackers including STRCF~\cite{Li18}, ECO~\cite{Danelljan16ECO}, SiamFC~\cite{Bertinetto16-2}, fDSST~\cite{Danelljan17}, Staple\_CA~\cite{Bertinetto16,Mueller17}, BT~\cite{Wang15} and MBT~\cite{Ma2016TIP}.  

\subsection{Comparative results}

 \begin{figure}[t]
	\begin{center}
		\includegraphics[width=0.98\linewidth]{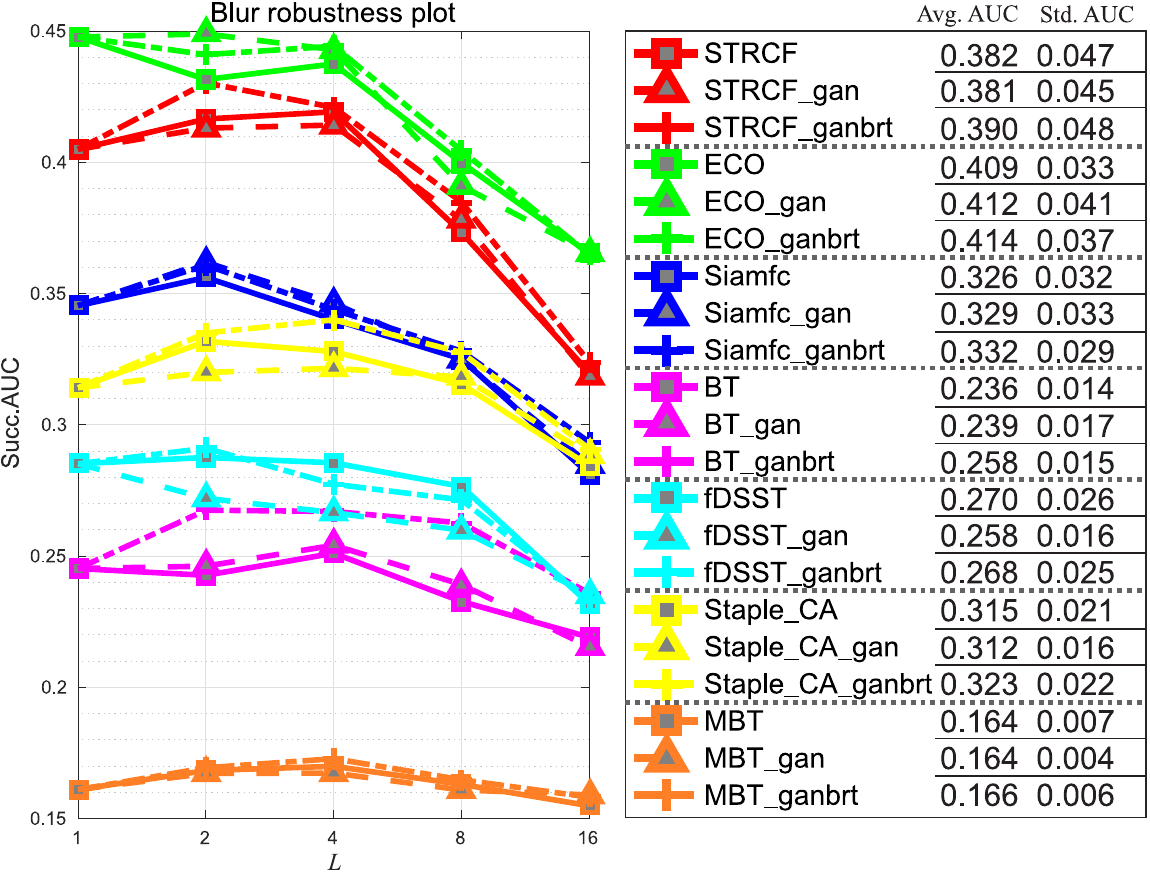}\vspace{-1em}
	\end{center}
	\caption{\small{
	Comparing proposed blur-robust trackers~(`*\_ganbrt') with full debluring~(`*\_gan') and non debluring~(`*') based trackers on $80\times 4$ blurred videos.  
	}}
	\label{fig:sltdeblur}\vspace{-1.1em}
\end{figure}

Since we have used 20 scenes, i.e. $20\times 4$ blurred videos, of the BVT benchmark to fine-tune the DeblurGAN-D in Section~\ref{subsec:fine-tuning}, the remaining 80 scenes form new subsets denoted as $\{\mathcal{S^{'}}^L|L=1,2,4,8,16\}$ each of them consists of 80 videos. We use them to validate the proposed blur robust tracking scheme on 7 trackers. 
We run the original trackers, full deburring~(`*\_gan') and the proposed scheme based versions~('*\_ganbrt') on $\{\mathcal{S^{'}}^L|L=2,4,8,16\}$ and calculate the average AUC and its standard variation as evaluation results.    
AUC scores of original trackers on $\mathcal{S^{'}}^1$ are also calculated for the comprehensive comparison.

As shown in Fig.~\ref{fig:sltdeblur}, according to the average AUCs, all trackers except fDSST are improved by the proposed blur-robust tracking scheme. In particular, BT\_ganbrt achieves 9.3\% relative improvement over the original version.
Moreover, the accuracy of BT\_ganbrt on $\mathcal{S^{'}}^{\{2,4,8\}}$ is much higher than the one of BT on the sharp subset, i.e. $\mathcal{S^{'}}^1$. 
STRCF\_ganbrt, ECO\_ganbrt, and Siamfc\_ganbrt outperforms STRCF, ECO and Siamfc on all subsets respectively. 
Staple\_CA\_ganbrt achieves 2.5\% relative improvement over Staple\_CA.
The accuracy increase of MBT\_ganbrt w.r.t. MBT is small since MBT is specifically designed for the tracking under motion blur. 
fDSST\_ganbrt obtains light worse accuracy than fDSST while being better on $\mathcal{S}^1$. 
More results are presented and discussed in the supplementary material.  
%
\section{Conclusion}

In this paper, we have proposed the Blurred Video Tracking~(BVT) benchmark to explore how motion blur affects visual object tracking and whether state-of-the-art deblurring can benefit the state-of-the-art trackers under different levels motion blur.
The proposed BVT benchmark contains 500 videos for 100 scenes, each of which has 5 videos with different levels of motion blurs. 
According to the evaluation results of 23 recent trackers on the BVT benchmark, we find that slight motion blur may have positive effects to visual tracking, while severe blurs certainly harm the performance of most trackers.
Using two state-of-the-art deblurring methods, DeblurGAN~\cite{Kupyn2018CVPR} and SRN~\cite{Tao2018CVPR}, to handle the blurred videos in our BVT benchmark, we study the effects of deblurring to 7 typical trackers. We  observe that current deblurring algorithm can improve tracking performance on severely blurred videos, while harm the accuracy on videos with slight motion blur.
Accordingly, we propose a general blur-robust tracking scheme that adopts a fine-tuned discriminator of DeblurGAN as an assessor to adaptively determine whether or not conduct deblurring for current frame.  
This method successfully improves the accuracy of 6 state-of-the-art trackers. In the future, we want to study how to generalize such adaptive deblurring strategy to further boost the robustness to blur in visual tracking.

{\small
\bibliographystyle{ieee}
\bibliography{IEEEabrv,deblurtrack}

\begin{thebibliography}{10}\itemsep=-1pt

\bibitem{Arjovsky2017}
M.~Arjovsky, S.~Chintala, and L.~Bottou.
\newblock Wasserstein gan.
\newblock {\em arXiv:1701.07875}, 2017.

\bibitem{Bao12}
C.~Bao, Y.~Wu, H.~Ling, and H.~Ji.
\newblock Real time robust l1 tracker using accelerated proximal gradient
  approach.
\newblock In {\em Proceedings of IEEE Conference on Computer Vision and Pattern
  Recognition}, pages 1830--1837, 2012.

\bibitem{Bertinetto16}
L.~Bertinetto, J.~Valmadre, S.~Golodetz, O.~Miksik, and P.~H.~S. Torr.
\newblock Staple: Complementary learners for real-time tracking.
\newblock In {\em Proceedings of IEEE Conference on Computer Vision and Pattern
  Recognition}, 2016.

\bibitem{Bertinetto16-2}
L.~Bertinetto, J.~Valmadre, J.~F. Henriques, A.~Vedaldi, and P.~H.~S. Torr.
\newblock Fully-convolutional siamese networks for object tracking.
\newblock In {\em arXiv preprint arXiv:1606.09549}, 2016.

\bibitem{Dai2006ICIP}
S.~Dai, M.~Yang, Y.~Wu, and A.~K. Katsaggelos.
\newblock Tracking motion-blurred targets in video.
\newblock In {\em Proceedings of IEEE International Conference on Image
  Processing}, 2006.

\bibitem{Danelljan16ECO}
M.~Danelljan, G.~Bhat, F.~S. Khan, and M.~Felsberg.
\newblock Eco: Efficient convolution operators for tracking.
\newblock In {\em Proceedings of IEEE Conference on Computer Vision and Pattern
  Recognition}, 2017.

\bibitem{Danelljan14}
M.~Danelljan, G.~H\"{a}ger, F.~S. Khan, and M.~Felsberg.
\newblock Accurate scale estimation for robust visual tracking.
\newblock In {\em Proceedings of the British Machine Vision Conference}, 2014.

\bibitem{Danelljan15}
M.~Danelljan, G.~H\"{a}ger, F.~S. Khan, and M.~Felsberg.
\newblock Learning spatially regularized correlation filters for visual
  tracking.
\newblock In {\em Proceedings of IEEE International Conference on Computer
  Vision}, pages 4310--4318, 2015.

\bibitem{Danelljan17}
M.~Danelljan, G.~H?ger, F.~S. Khan, and M.~Felsberg.
\newblock Discriminative scale space tracking.
\newblock {\em {IEEE} Transactions on Pattern Analysis and Machine
  Intelligence}, 39(8):1561--1575, 2018.

\bibitem{Ding2016TCSVT}
J.~{Ding}, Y.~{Huang}, W.~{Liu}, and K.~{Huang}.
\newblock Severely blurred object tracking by learning deep image
  representations.
\newblock {\em IEEE Transactions on Circuits and Systems for Video Technology},
  26(2):319--331, 2016.

\bibitem{Dolhansky_2018_CVPR}
B.~Dolhansky and C.~Canton~Ferrer.
\newblock Eye in-painting with exemplar generative adversarial networks.
\newblock In {\em Proceedings of IEEE Conference on Computer Vision and Pattern
  Recognition}, 2018.

\bibitem{Fan2019LaSOT}
H.~Fan, L.~Lin, F.~Yang, P.~Chu, G.~Deng, S.~Yu, H.~Bai, Y.~Xu, C.~Liao, and
  H.~Ling.
\newblock Lasot: A high-quality benchmark for large-scale single object
  tracking.
\newblock In {\em Proceedings of IEEE Conference on Computer Vision and Pattern
  Recognition}, 2019.

\bibitem{Fan2019CVPR}
H.~Fan and H.~Ling.
\newblock Siamese cascaded region proposal networks for real-time visual
  tracking.
\newblock In {\em Proceedings of IEEE Conference on Computer Vision and Pattern
  Recognition}, 2019.

\bibitem{DSARCF_TIP2019}
W.~Feng, R.~Han, Q.~Guo, J.~K. Zhu, and S.~Wang.
\newblock Dynamic saliency-aware regularization for correlation filter based
  object tracking.
\newblock {\em {IEEE} Transactions on Image Processing}, pages 1--1, 2019.

\bibitem{Galoogahi2017ICCV}
H.~K. Galoogahi, A.~Fagg, C.~Huang, D.~Ramanan, and S.~Lucey.
\newblock Need for speed: A benchmark for higher frame rate object tracking.
\newblock In {\em Proceedings of IEEE International Conference on Computer
  Vision}, 2017.

\bibitem{Galoogahi17}
H.~K. Galoogahi, A.~Fagg, and S.~Lucey.
\newblock Learning background-aware correlation filters for visual tracking.
\newblock In {\em Proceedings of IEEE International Conference on Computer
  Vision}, pages 1144--1152, 2017.

\bibitem{Goodfellow2014}
I.~J. Goodfellow, J.~Pouget-Abadie, M.~Mirza, B.~Xu, D.Warde-Farley, S.~Ozair,
  A.~Courville, and Y.~Bengio.
\newblock Generative adversarial networks.
\newblock 2014.

\bibitem{Gulrajani2017}
I.~Gulrajani, F.~Ahmed, M.~Arjovsky, V.~Dumoulin, and A.~Courville.
\newblock Improved training of wasserstein gans.
\newblock {\em arXiv:1704.00028}, 2017.

\bibitem{Guo17_ICCV}
Q.~Guo, W.~Feng, C.~Zhou, R.~Huang, L.~Wan, and S.~Wang.
\newblock Learning dynamic {Siamese} network for visual object tracking.
\newblock In {\em Proceedings of IEEE International Conference on Computer
  Vision}, 2017.

\bibitem{Henriques12}
J.~F. Henriques, R.~Caseiro, P.~Martins, and J.~Batista.
\newblock Exploiting the circulant structure of tracking-by-detection with
  kernels.
\newblock In {\em Proceedings of the European Conference on Computer Vision},
  pages 702--715, 2012.

\bibitem{Henriques15}
J.~F. Henriques, R.~Caseiro, P.~Martins, and J.~Batista.
\newblock High-speed tracking with kernelized correlation filters.
\newblock {\em {IEEE} Transactions on Pattern Analysis and Machine
  Intelligence}, 37(3):583--596, 2015.

\bibitem{huang2018got}
L.~Huang, X.~Zhao, and K.~Huang.
\newblock Got-10k: A large high-diversity benchmark for generic object tracking
  in the wild.
\newblock {\em arXiv preprint arXiv:1810.11981}, 2018.

\bibitem{Jin2005CVPR}
H.~Jin, P.~Favaro, and R.~Cipolla.
\newblock Visual tracking in the presence of motion blur.
\newblock volume~2, pages 18--25, 2005.

\bibitem{Jung2018ECCV}
I.~Jung, J.~Son, M.~Baek, and B.~Han.
\newblock Real-time mdnet.
\newblock In {\em Proceedings of the European Conference on Computer Vision},
  2018.

\bibitem{Kristan2017ICCVW}
M.~Kristan, A.~Leonardis, J.~Matas, M.~Felsberg, R.~Pflugfelder, L.~\v{C}ehovin
  Zajc, T.~Vojir, G.~H\"{a}ger, A.~Luke\v{z}i\v{c}, and G.~Fernandez.
\newblock The visual object tracking vot2017 challenge results.
\newblock In {\em Proceedings of IEEE International Conference on Computer
  Vision Workshop}, 2017.

\bibitem{Kristan15}
M.~Kristan, J.~Matas, A.~Leonardis, M.~Felsberg, L.~Cehovin, G.~Fernandez,
  T.~Vojir, G.~Hager, G.~Nebehay, R.~Pflugfelder, A.~Gupta, A.~Bibi,
  A.~Lukezic, A.~Garcia-Martin, A.~Saffari, A.~Petrosino, and A.~S. Montero.
\newblock The visual object tracking {VOT2015} challenge results.
\newblock In {\em Proceedings of IEEE International Conference on Computer
  Vision Workshop}, pages 564--586, 2015.

\bibitem{Kristan16}
M.~Kristan, J.~Matas, A.~Leonardis, T.~Vojir, R.~Pflugfelder, G.~Fernandez,
  G.~Nebehay, F.~Porikli, and L.~Cehovin.
\newblock A novel performance evaluation methodology for single-target
  trackers.
\newblock {\em {IEEE} Transactions on Pattern Analysis and Machine
  Intelligence}, 38(11):2137--2155, 2016.

\bibitem{Kupyn2018CVPR}
O.~Kupyn, V.~Budzan, M.~Mykhailych1, D.~Mishkin, and J.~Matas.
\newblock Deblurgan:blind motion deblurring using conditional adversarial
  networks.
\newblock In {\em Proceedings of IEEE Conference on Computer Vision and Pattern
  Recognition}, 2018.

\bibitem{Ledig2017CVPR}
C.~{Ledig}, L.~{Theis}, F.~{Huszár}, J.~{Caballero}, A.~{Cunningham},
  A.~{Acosta}, A.~{Aitken}, A.~{Tejani}, J.~{Totz}, Z.~{Wang}, and W.~{Shi}.
\newblock Photo-realistic single image super-resolution using a generative
  adversarial network.
\newblock In {\em Proceedings of IEEE Conference on Computer Vision and Pattern
  Recognition}, pages 105--114, 2017.

\bibitem{Li2016PAMI}
A.~Li, M.~Lin, Y.~Wu, M.-H. Yang, and S.~Yan.
\newblock Nus-pro:a new visual tracking challenge.
\newblock {\em {IEEE} Transactions on Pattern Analysis and Machine
  Intelligence}, 38(2):335--349, 2016.

\bibitem{Li18}
F.~Li, C.~Tian, W.~Zuo, L.~Zhang, and M.-H. Yang.
\newblock Learning spatial-temporal regularized correlation filters for visual
  tracking.
\newblock In {\em Proceedings of IEEE Conference on Computer Vision and Pattern
  Recognition}, pages 4904--4913, 2018.

\bibitem{Li14-samf}
Y.~Li and J.~Zhu.
\newblock A scale adaptive kernel correlation filter tracker with feature
  integration.
\newblock In {\em Proceedings of the European Conference on Computer Vision
  Workshop}, 2014.

\bibitem{Liang2015TIP}
P.~Liang, E.~Blasch, and H.~Ling.
\newblock Encoding color information for visual tracking: Algorithms and
  benchmark.
\newblock {\em {IEEE} Transactions on Image Processing}, 24(12):5630--5644,
  2015.

\bibitem{AL17-csrdcf}
A.~Luke\v{z}i\v{c}, T.~Voj\'{i}\v{r}, L.~\v{C}ehovin, J.~Matas, and M.~Kristan.
\newblock Discriminative correlation filter with channel and spatial
  reliability.
\newblock In {\em Proceedings of IEEE Conference on Computer Vision and Pattern
  Recognition}, 2017.

\bibitem{Ma2016TIP}
B.~Ma, L.~Huang, J.~Shen, L.~Shao, M.-H. Yang, and F.~Porikli.
\newblock Visual tracking under motion blur.
\newblock {\em {IEEE} Transactions on Image Processing}, 25(12):5867--5876,
  2016.

\bibitem{Ma15}
C.~Ma, J.~B. Huang, X.~Yang, and M.~H. Yang.
\newblock Hierarchical convolutional features for visual tracking.
\newblock In {\em Proceedings of IEEE International Conference on Computer
  Vision}, 2015.

\bibitem{Mueller17}
N.~S. Matthias~Mueller and B.~Ghanem.
\newblock Context-aware correlation filter tracking.
\newblock In {\em Proceedings of IEEE Conference on Computer Vision and Pattern
  Recognition}, 2017.

\bibitem{Mei2008CVPR}
C.~{Mei} and I.~{Reid}.
\newblock Modeling and generating complex motion blur for real-time tracking.
\newblock In {\em Proceedings of IEEE Conference on Computer Vision and Pattern
  Recognition}, pages 1--8, 2008.

\bibitem{Mueller2018ECCV}
M.~Mueller, A.~Bibi, S.~Giancola, S.~Al-Subaihi, and B.~Ghanem.
\newblock Trackingnet: A large-scale dataset and benchmark for object tracking
  in the wild.
\newblock In {\em Proceedings of the European Conference on Computer Vision},
  2018.

\bibitem{Mueller2016ECCV}
M.~Mueller, N.~Smith, and B.~Ghanem.
\newblock A benchmark and simulator for uav trackings.
\newblock In {\em Proceedings of the European Conference on Computer Vision},
  2016.

\bibitem{Nah2017CVPR}
S.~Nah, T.~H. Kim, and K.~M. Lee.
\newblock Deep multi-scale convolutional neural network for dynamic scene
  deblurring.
\newblock In {\em Proceedings of IEEE Conference on Computer Vision and Pattern
  Recognition}, 2017.

\bibitem{MDNet16}
H.~Nam and B.~Han.
\newblock Learning multi-domain convolutional neural networks for visual
  tracking.
\newblock In {\em Proceedings of IEEE Conference on Computer Vision and Pattern
  Recognition}, pages 4293--4302, 2016.

\bibitem{Nam16}
H.~Nam and B.~Han.
\newblock Learning multi-domain convolutional neural networks for visual
  tracking.
\newblock In {\em Proceedings of IEEE Conference on Computer Vision and Pattern
  Recognition}, 2016.

\bibitem{Noroozi2017}
M.~Noroozi, P.~Chandramouli, and P.~Favaro.
\newblock Motion deblurring in the wild.
\newblock In {\em Pattern Recognition}, pages 65--77, 2017.

\bibitem{Ross07}
D.~A. Ross, J.~Lim, R.-S. Lin, and M.-H. Yang.
\newblock Incremental learning for robust visual tracking.
\newblock {\em International Journal of Computer Vision}, 77(1):125--141, 2007.

\bibitem{Salimans2016}
T.~Salimans, I.~Goodfellow, W.~Zaremba, V.~Cheung, A.~Radford, , and X.~Chen.
\newblock Improved techniques for training gans.
\newblock {\em arXiv:1606.03498}, 2016.

\bibitem{Seibold2017CVIU}
C.~Seibold, A.~Hilsmann, and P.~Eisert.
\newblock Model-based motion blur estimation for the improvement of motion
  tracking.
\newblock {\em Computer Vision and Image Understanding}, 160:45--56, 2017.

\bibitem{Smeulders2014PAMI}
A.~W.~M. {Smeulders}, D.~M. {Chu}, R.~{Cucchiara}, S.~{Calderara},
  A.~{Dehghan}, and M.~{Shah}.
\newblock Visual tracking: An experimental survey.
\newblock {\em IEEE Transactions on Pattern Analysis and Machine Intelligence},
  36(7):1442--1468, 2014.

\bibitem{Song2018CVPR}
Y.~Song, C.~Ma, X.~Wu, L.~Gong, L.~Bao, W.~Zuo, C.~Shen, R.~W. Lau, and M.-H.
  Yang.
\newblock Vital:visual tracking via adversarial learning.
\newblock In {\em Proceedings of IEEE Conference on Computer Vision and Pattern
  Recognition}, 2018.

\bibitem{Sun2015CVPR}
J.~Sun, W.~Cao, Z.~Xu, and J.~Ponce.
\newblock Learning a convolutional neural network for non-uniform motion blur
  removal.
\newblock In {\em Proceedings of IEEE Conference on Computer Vision and Pattern
  Recognition}, 2015.

\bibitem{Tao2018CVPR}
X.~Tao, H.~Gao, X.~Shen, J.~Wang, and J.~Jia.
\newblock Scale-recurrent network for deep image deblurring.
\newblock In {\em Proceedings of IEEE Conference on Computer Vision and Pattern
  Recognition}, 2018.

\bibitem{Wang15}
N.~Wang, J.~Shi, D.~Y. Yeung, and J.~Jia.
\newblock Understanding and diagnosing visual tracking systems.
\newblock In {\em Proceedings of IEEE International Conference on Computer
  Vision}, pages 3101--3109, 2015.

\bibitem{Wang2019CVPR}
Q.~Wang, L.~Zhang, L.~Bertinetto, W.~Hu, and P.~H. Torr.
\newblock Fast online object tracking and segmentation: A unifying approach.
\newblock In {\em Proceedings of IEEE Conference on Computer Vision and Pattern
  Recognition}, 2019.

\bibitem{Wang2018CVPR}
X.~Wang, C.~Li, B.~Luo, and J.~Tang.
\newblock Sint++:robust visual tracking via adversarial positive instance
  generation.
\newblock In {\em Proceedings of IEEE Conference on Computer Vision and Pattern
  Recognition}, 2018.

\bibitem{Wu2011ICCV}
Y.~{Wu}, , , , and and.
\newblock Blurred target tracking by blur-driven tracker.
\newblock In {\em Proceedings of IEEE International Conference on Computer
  Vision}, pages 1100--1107, 2011.

\bibitem{Wu2011AVC}
Y.~Wu, J.~Hu, F.~Li, E.~Cheng, J.~Yu, and H.~Ling.
\newblock Kernel-based motion-blurred target tracking.
\newblock In {\em Advances in Visual Computing}, pages 486--495, 2011.

\bibitem{Wu13}
Y.~Wu, J.~Lim, and M.-H. Yang.
\newblock Online object tracking: a benchmark.
\newblock In {\em Proceedings of IEEE Conference on Computer Vision and Pattern
  Recognition}, pages 2411--2418, 2013.

\bibitem{Wu15}
Y.~Wu, J.~Lim, and M.-H. Yang.
\newblock Object tracking benchmark.
\newblock {\em {IEEE} Transactions on Pattern Analysis and Machine
  Intelligence}, 37(9):1834--1848, 2015.

\bibitem{Xu2014NIPS}
L.~Xu, J.~S.~J. Ren, C.~Liu, and J.~Jia.
\newblock Deep convolutional neural network for image deconvolution.
\newblock pages 1790--1798, 2014.

\bibitem{Yeh2017CVPR}
R.~A. {Yeh}, C.~{Chen}, T.~Y. {Lim}, A.~G. {Schwing}, M.~{Hasegawa-Johnson},
  and M.~N. {Do}.
\newblock Semantic image inpainting with deep generative models.
\newblock In {\em Proceedings of IEEE Conference on Computer Vision and Pattern
  Recognition}, pages 6882--6890, 2017.

\bibitem{Zhang12ECCV}
K.~Zhang, L.~Zhang, and M.-H. Yang.
\newblock Real-time compressive tracking.
\newblock In {\em Proceedings of the European Conference on Computer Vision},
  volume~36, page 864–877, 2012.

\bibitem{Zhang14-stc}
K.~Zhang, L.~Zhang, M.-H. Yang, and D.~Zhang.
\newblock Fast tracking via dense spatio-temporal context learning.
\newblock In {\em Proceedings of the European Conference on Computer Vision},
  pages 127--141, 2014.

\bibitem{Zhu2018ECCV}
Z.~Zhu, Q.~Wang, B.~Li, W.~Wei, J.~Yan, and W.~Hu.
\newblock Distractor-aware siamese networks for visual object tracking.
\newblock In {\em Proceedings of the European Conference on Computer Vision},
  2018.

\end{thebibliography}
}

\end{document}